%% file: main.tex
\documentclass{article}
\usepackage{spconf,amsmath,graphicx,hyperref}
\usepackage{cite}
\usepackage{amssymb}
\usepackage{amsmath}
\usepackage{algorithmic}
\usepackage{multirow}
\usepackage{booktabs}
\usepackage{colortbl, xcolor}
\usepackage{bm}
\usepackage{float} % preamble

\title{MedCutMix: A Data-Centric Approach to
Improve Radiology Vision-Language Pre-training
with Disease Awareness}
\name{Sinuo Wang$^1$ \qquad Yutong Xie$^2$ \qquad Yuyuan Liu$^3$ \qquad Qi Wu$^1$}
  
\address{$^1$The University of Adelaide \\
$^2$Mohamed bin Zayed University of Artificial Intelligence  \\ $^3$University of Oxford}
      
\begin{document}
%\ninept
%
\maketitle

\small % 9pt body

\begin{abstract}
Vision-Language Pre-training (VLP) is drawing increasing interest for its ability to minimize manual annotation requirements while enhancing semantic understanding in downstream tasks. However, its reliance on image-text datasets poses challenges due to privacy concerns and the high cost of obtaining paired annotations.
Data augmentation emerges as a viable strategy to address this issue, yet existing methods often fall short of capturing the subtle and complex variations in medical data due to limited diversity. 
To this end, we propose MedCutMix, a novel multi-modal disease-centric data augmentation method. MedCutMix performs diagnostic sentence CutMix within medical reports and establishes the cross-attention between the diagnostic sentence and medical image to guide attentive manifold mix within the imaging modality. Our approach surpasses previous methods across four downstream radiology diagnosis datasets, highlighting its effectiveness in enhancing performance and generalizability in radiology VLP.
\end{abstract}
\begin{keywords}
Medical VLP, Disease-aware Augmentation
\end{keywords}
\section{Introduction}
\label{sec:intro}

% medical VLP 重要
% medical VLP的进展：gobal alignment --> local alignment: MGCA 但是仍然面临数据量有限的挑战
% 我们提出Data-Centric medical VLP方法，通过扩充医学数据对，从数据增广的角度来改善medical VLP的性能。现有的数据增广在medical VLP里面的challenge（参考PairAug）
% 我们提出MedCutMix方法，是CutMix在medical VLP的重要进展。CutMix是先进的数据增广方法，它的的原理是，需要强的分类信号获得图像重要区域的定位信息，进行mix。但是没有研究将CutMix用于medical VLP，因为在medical VLP获得定位信息是比较难的。
% 我们是怎么把CutMix用进来的，首先fine-grained medical VLP进行image-text的匹配，提出disease-prompt，找到定位信息，进行feature-level mixup (为什么不是image-level mixup，时间和memory).

The rise of self-supervised deep learning methods \cite{clip_radford2021learning, albef_li2021align, blip_li2022blip} has attracted significant attention due to their capacity to reduce the need for extensive manual annotations. This advantage is especially pronounced in the medical field, where the annotation process is costly but the multimodal nature of medical data often allows weak supervision through accompanying medical reports \cite{convirt_zhang2022contrastive, mrm_zhou2023advancing, ptunifier_chen2023towards}.
Leveraging the paired image-text data, the Vision-Language Pre-training (VLP) paradigm enables the model to learn general visual and textual representations, offering data-efficient solutions for downstream target tasks. In contrast to the general domain, medical VLP models necessitate exceeding global alignment and acquiring the ability to capture fine-grained representations. In radiology, clinicians often focus on the subtle yet crucial visual cues to perform the diagnosis, which cannot be adequately captured solely through the global alignment objective \cite{medclip_wang2022medclip, gloria_huang2021gloria, mgca_wang2022multi, shui2025largescalefinegrainedvisionlanguagepretraining, lovt_muller2022joint}. To address this, recent works propose attention-based local alignment \cite{gloria_huang2021gloria} and disease-level alignment strategies \cite{mgca_wang2022multi}. In the masked image modeling pretraining, MedIM \cite{medim_xie2023medim} proposes masking and reconstructing the image regions guided by reports to enhance the fine-grained semantics capturing. 

Despite these advancements, these approaches persist in their inherent data-hunger nature and face the challenge of medical data scarcity. While widely adopted VLP models in general domains, such as CLIP \cite{clip_radford2021learning}, are trained on vast datasets of 400M image-text pairs from the internet, replicating such efforts for medical data is currently hindered by privacy and legal concerns. 
We advocate a data-centric approach to enhance medical VLP by focusing on data augmentation to expand medical image-report pairs. It offers a promising solution to mitigate data scarcity constraints while simultaneously enriching data diversity without requiring real-world data acquisition, thus maintaining patient privacy.

% Many studies~\cite{zhao2019data,garcea2023data} have explored traditional spatial transformations and morphological operations to improve model generalization. However, these methods face limitations in effectively augmenting medical images while ensuring that the modifications remain semantically aligned with their corresponding text. 
% An alternative is Mixup \cite{mixup_zhang2017mixup}, a popular data augmentation technique, that linearly blends images and their labels from two different categories to generate new training samples~\cite{mixup_survey_jin2024survey}. Mixup has shown effectiveness in enhancing the model's ability to generalize across medical domains \cite{galdran2021balanced,gong2022vqamix}. However, when applied to medical VLP tasks, such global linear mixing approaches may diminish the focus on subtle yet critical visual cues.
% % 
% In contrast, CutMix~\cite{cutmix_yun2019cutmix, attentive_cutmix_walawalkar2020attentive} offers a locality-aware augmentation strategy that simultaneously mixes images and labels, preserving disease-specific regions while maintaining label integrity. However, its potential application in medical VLP remains largely unexplored due to two key challenges: first, accurately extracting disease information from medical reports, which typically contain complex terminology; second, obtaining image localization information for specific diseases in VLP tasks without explicit annotations. 

Recognizing this gap, we propose MedCutMix, a novel multi-modal disease-centric data augmentation technique explicitly designed for medical vision-language pre-training. MedCutMix not only increases the volume of the training data for medical VLP but also expands the data diversity, specifically focusing on subtle yet crucial disease-related semantics multi-modally. 
% 
% Specifically, we leverage the disease labels and pathological information available in each report to identify diagnostic sentences. We then perform input-level CutMix on these sentences between the source and target reports and use the diagnostic sentences to guide image attention. Cross-attention is employed to conduct an attentive mix for images at the feature level. 
To achieve this, we extract disease labels from radiology reports using the rule-based labeler. These labels help identify diagnostic sentences, which contain key medical findings relevant to the disease. 
To effectively integrate this textual information with images, 
we introduce a pairwise CutMix strategy: 
\textit{(1) Text-level CutMix}:
% \yy{Should we avoid the CutMix?}
% \yy{our augmentation strategy is applied for both modalities: \textit{(1) text injection}: }
diagnostic sentences from source and target reports are mixed at the input level to augment textual diversity; and \textit{(2) Feature-level Image Mixing}: we leverage cross-attention between diagnostic sentences and image features to identify disease-relevant image regions, which are then selectively mixed at the feature level, ensuring semantic consistency between the mixed image and mixed report. Our contributions include: 
\begin{enumerate}
\item We propose MedCutMix, a novel disease-centric data augmentation framework designed for medical VLP, enhancing training data diversity while preserving critical disease-related semantics.
\item To ensure cross-modal consistency, we introduce a pairwise CutMix strategy that integrates text-level CutMix for augmenting diagnostic sentences and feature-level image mixing, where cross-attention aligns disease-relevant visual regions with textual information.
\item 
% \ul{We conducted medical VLP experiments on the widely used MIMIC-CXR dataset, comparing the performance with and without MedCutMix. Comprehensive evaluations across four downstream radiology diagnosis datasets show that MedCutMix significantly outperforms the advanced baseline, demonstrating its effectiveness in improving model performance and generalizability in radiology VLP.}
Our results show that MedCutMix successfully improves performance over previous methods across four downstream radiology diagnosis datasets, demonstrating its effectiveness in enhancing generalisation in radiology VLP.
\end{enumerate}
\vspace{-10pt}

\section{Method}
\label{sec:Method}
\begin{figure*}[t!]
\centering
 \includegraphics[width=\textwidth]{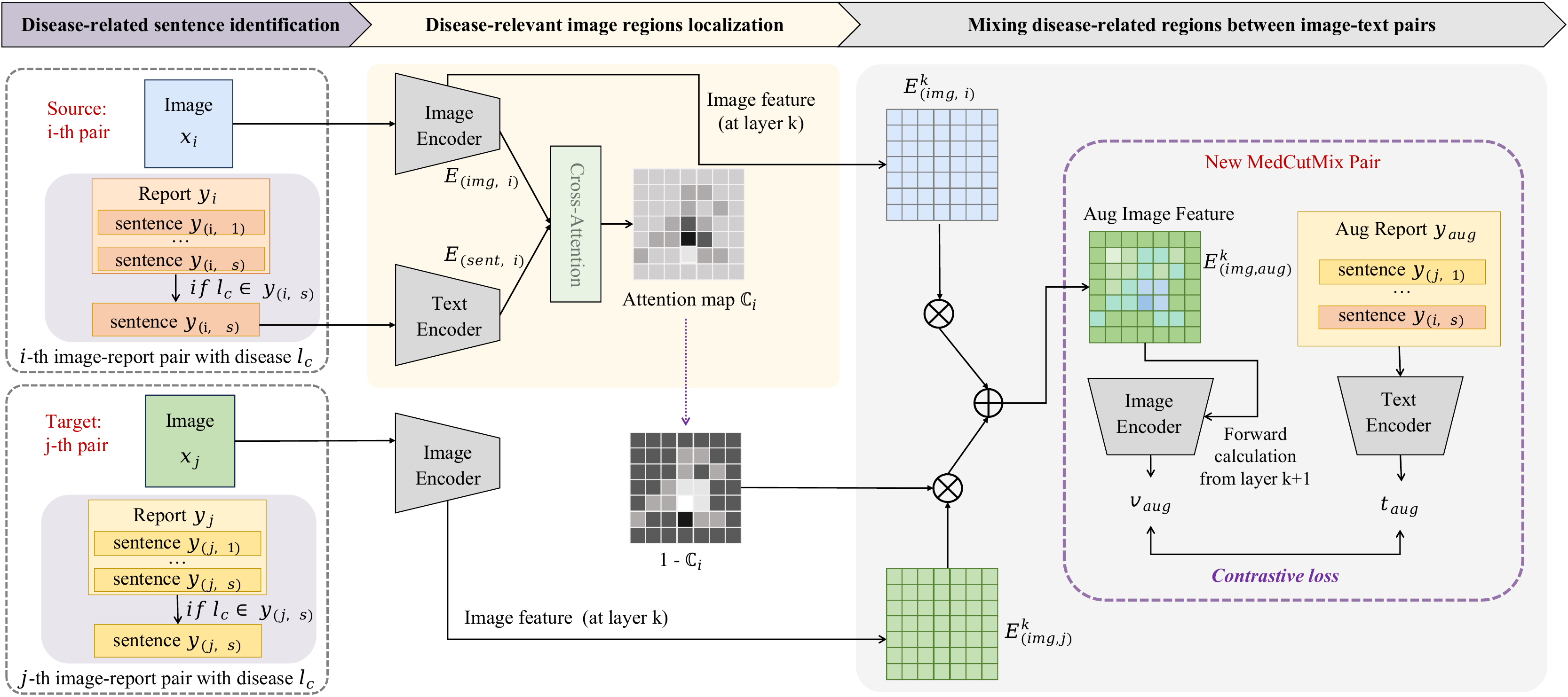}
  \vspace{-20pt}
  \caption{Overview of the MedCutMix pipeline. After disease-centric label extraction and balanced sampling, the pipeline proceeds with the following steps: (1) identification of disease-related sentences from medical reports, (2) localization of disease-relevant regions in medical images, and (3) mixing disease-related regions between image-text pairs to maintain semantic consistency.}
    \vspace{-10pt}
\label{fig1}
\end{figure*}

In this work, we propose MedCutMix, a multi-modal disease-centric data augmentation method tailored for medical VLP. As shown in Fig.~\ref{fig1}, MedCutMix introduces a novel pairwise CutMix strategy, transferring disease-related regions across image-report pairs to enhance the model’s ability to recognize and understand diseases in diverse visual contexts. The method comprises four key steps: (1) Disease-centric label extraction and balanced sampling, (2) Disease-related sentence identification from medical reports, (3) Disease-relevant regions localization from medical images, and (4) Mixing disease-related regions between image-text pairs to maintain 
semantic consistency.\\
% An effective data augmentation for medical VLP should meet two key criteria: (1) disease-related regions should be extracted and transferred to enhance semantic information, and (2) the cutout regions between images and reports must maintain semantic alignment.

\vspace{-0.2cm}
\noindent \textbf{Disease-centric label extraction and balanced sampling.} In this work, we use the MIMIC-CXR dataset~\cite{mimiccxr_johnson2019mimic} and focus on $C$ common chest X-ray disease findings for pairwise CutMix. 
% We extract the disease label from MIMIC-CXR reports via the open-source rule-based tool-CheXpert labeler~\cite{cxp_irvin2019chexpert} and denote the label set as $\{l_{c}\}_{c=1}^C$.
% 
Let the pre-training dataset be denoted by $\mathcal{D} = \left\{(x_i, y_i, l_{(i,c)}) \mid c = 1, 2, ..., C  )\right\}^{|D|}_{i=1}$, where \(x_i \in \mathcal{X} \in \mathbb{R}^{H \times W}\) represents the input image resolution with 3 RGB channels, and \(y_i \in \mathcal{Y} \in \mathbb{R}^{1 \times N_{si}}\) denotes the \(N_{si}\) sentences in each report that describe the corresponding image \(x_i\). 
The disease label \(l_{(i,c)}\) is extracted from MIMIC-CXR reports via the open-source rule-based tool-CheXpert labeler~\cite{cxp_irvin2019chexpert}, where \(C\) denotes the total categories of the dataset.
To address the class imbalance, we perform balanced sampling by selecting $n = \frac{N_{max}}{c}$ instances for each disease, where $N_{max}$ is a hyperparameter specifying the maximum number of newly created pairs.
For each pair of randomly sampled image-report pairs, e.g., $(\bm{x}_i,\bm{y}_i)$ and $(\bm{x}_j,\bm{y}_j)$, %from Fig.~\ref{fig1}, 
both of which contain disease $l_{c}$, we identify diagnostic sentences and visual regions pertinent to disease $l_{c}$. We then perform input-level CutMix for the medical report and execute attentive feature-level mixup for the corresponding image modality.\\

\vspace{-0.2cm}
\noindent \textbf{Disease-related sentence identification from medical reports.}
Recognizing that radiological reports often comprise several sentences, each potentially providing independent information about different findings of the image~\cite{biovil_boecking2022making}. Diagnostic sentences not only identify the disease but also convey additional details such as severity and other relevant clinical aspects, thus offering a more comprehensive understanding than a mere mention of disease name. Consequently, we extract disease-related content from medical reports at the sentence level. Specifically, we perform the semantic match for the disease $l_{c}$ within different sentences in $\bm{y}_i= \{\bm{y}_{(i,s)}\}_{s=1}^{N_{si}}$ and $\bm{y}_j= \{\bm{y}_{(j,s)}\}_{s=1}^{N_{sj}}$, where $N_{si}$ and $N_{sj}$ are the number of sentences. 
The sentence containing an exact match of the disease term \( l_c \) is cut out as the diagnostic sentence, denoted as \(y_{(i,s)}^{l_c} = {\rm \texttt{Cutout}}(\bm{y}_i, M_{ti})\) and \( y_{(j,s)}^{l_c} = {\rm \texttt{Cutout}}(\bm{y}_j, M_{tj}) \). Here, $M_{ti}$ and $M_{tj}$ are the binary masks that point out where the disease-related sentence in the reports, as follows:
\begin{equation}
\label{eq:IntraTextMask}
\begin{aligned}
M_{ti}=\{M_{(ti, s)}=\mathbb{I}(l_{c} \in \bm{y}_{(i,s)})\} \in \{0, 1\}^{N_{si}}, \\
M_{tj}=\{M_{(tj, s)}=\mathbb{I}(l_{c} \in \bm{y}_{(j,s)})\} \in \{0, 1\}^{N_{sj}},
\end{aligned}
\end{equation}
where \( \mathbb{I} \) is an indicator function that assigns \( 1 \) to sentences containing \( l_c \) and \( 0 \) otherwise. The operation \({\rm \texttt{Cutout}(\cdot, \cdot)}\) selects a sentence from \(y_i\) or \(y_j\) where the corresponding binary mask \(M_{(ti,s)} = 1\) or \(M_{(tj,s)} = 1\). 

\vspace{+0.2cm}
\noindent \textbf{Disease-relevant regions localization from medical images.}
To achieve the disease semantic consistency across the augmented image and corresponding report, we guide the extraction of image regions using the diagnostic sentences. This process utilizes a warmed-up medical VLP model~\cite{mgca_wang2022multi} with the Vision Transformer~\cite{vaswani2017attention}, ViT-B/16, as image encoder, which aligns the fine-grained correspondences between medical images' visual and semantic aspects with their corresponding radiological reports, benefiting in better identifying the text-guided discriminative image regions.
The base
% pre-trained medical VLP 
model accepts images $\bm{x}_i$ and $\bm{x}_j$, along with their paired reports $\bm{y}_i$ and $\bm{y}_j$ as inputs. The model generates the image global embedding $\bm{v}_{i}$ and $\bm{v}_{j}$, local patch embeddings $\bm{E}_{(img, i)}$ and $\bm{E}_{(img, j)}$, as well as the global text representations $\bm{t}_{i}$ and $\bm{t}_{j}$, local embeddings $\bm{E}_{(text, i)}$ and $\bm{E}_{(text, j)}$.
% We also collect the intermediate image global representaions $\bm{v}_{i}^k$ and $\bm{v}_{j}^k$ and local features $\bm{E}_{(img, i)}^k$ and $\bm{E}_{(img, j)}^k$ from the $k$-th layer from the image encoder for future image feature-level mixup. 
%
% Unlike text, disease information within an image is not organized in a structured manner in the diagnostic sentence; the visual disease information is often dispersed in the image space. Consequently, to augment the image modality, we employ feature-level mixing to perform soft mixing at multiple locations simultaneously. 
%
We also collect the intermediate image local features $\bm{E}_{(img, i)}^k$ and $\bm{E}_{(img, j)}^k$ from the $k$-th layer from the image encoder for future image feature-level mixup. 
To target on the $l_{c}$ disease-related information, the diagnostic sentence embedding is derived by applying a masking operation over the report, represented as $\bm{E}_{(sent, i)} = {\texttt{Cutout}(\bm{E}_{(text, i)}, M_{ti})}$. Leveraging the intrinsic vision-language alignment capability of the warmed-up medical VLP model, the disease-related image regions are identified using the diagnostic sentence embeddings to attend to the image local embeddings. 
Specifically, we first compute the attention map $\mathbb{C}$ for the $i$-th pair as follows:
\begin{equation}
\begin{split}
\mathbb{C}_i = \sum \mathrm{softmax} (\frac{\bm{E}_{(img, i)} \cdot (\bm{E}_{(sent, i)})^{\top}}{\mathbb{\tau}_{1}}) \in \mathbb{R}^{N_{patch}}.
\end{split}
\end{equation}
Here, $N_{patch}$ denotes the number of image tokens obtained after flattening the 2D patchified image. The softmax function normalizes the elements along the image dimension to find the focused region matched to each word in the diagnostic sentence. The summation operation $\sum$ performs on the text dimension to aggregate the attention related to the whole diagnostic sentence. $\mathbb{\tau}$ is the temperature to control the size of the attention area. 

\vspace{+0.2cm}
\noindent \textbf{Mixing disease-related regions between image-text pairs.}
After identifying the disease-related sentences and their related image regions, we implement the mix augmentation. 
This text augmentation process operates as an input-level CutMix, involving cutting out the diagnostic sentence from the source report $\bm{y}_i$ and pasting the diagnostic sentence to the target report $\bm{y}_j$, represented as: 
\begin{equation}
\bm{y}_{aug} =  {\rm { \texttt{Paste} \Big{(}\texttt{Cutout}}}(\bm{y}_i, M_{ti}), \bm{y}_j, M_{tj} \Big{)},
\end{equation}
where \( \texttt{Paste} (\cdot,\cdot,\cdot)\) denotes the operation of placing the coutout from $\bm{y}_i$ into $\bm{y}_j$ at the location specified by the mask $M_{tj}$.
For the visual modality, we utilize the source attention $\mathbb{C}_i$  to perform an attentive soft mix on the intermediate features obtained from the $k$-th layer of the image encoder, expressed as: 
\begin{equation}
\bm{E}_{(img, aug)}^k = \mathbb{C}_i \odot \bm{E}_{(img, i)}^k  + (1-\mathbb{C}_i) \odot \bm{E}_{(img, j)}^k.
\end{equation}
Following this, the augmented report $\bm{y}_{aug}$ is encoded via the text encoder to yield the global text representation, denoted as $\bm{t}_{aug}$. whereas, the augmented image feature $\bm{E}_{(img, aug)}^k$ is directly fed back into the image encoder to resume encoding from layer $k+1$ onward, and derive the global visual representation $\bm{v}_{aug}$.
% In contrast, for the visual modality, we reconstruct the visual token sequence at layer $k$ by concatenating the original target global visual representation $\bm{v}_{j}^k$ and mixed visual patch feature $\bm{E}_{img, aug}^k$, and utilize the vision encoder to continue inference from layer $k+1 $ onward to obtain the global visual representation $\bm{v}_{aug}$.
The newly augmented global representations are subsequently used to compute another image-text contrastive loss, as follows:
\begin{equation}
\begin{aligned}
\ell_{(aug,i)}^{v2t} &= - \log \frac{\exp \left(\text{sim}\left(\bm{v}_{(aug, i)}, \bm{t}_{(aug, i)}\right) / \tau_2\right)}
{\sum_{j=1}^N \exp \left(\text{sim}\left(\bm{v}_{(aug, i)}, \bm{t}_{(aug, j)}\right) / \tau_2\right)}, \\
\ell_{(aug, i)}^{t2v} &= - \log \frac{\exp \left(\text{sim}\left(\bm{t}_{(aug, i)}, \bm{v}_{(aug, i)}\right) / \tau_2\right)}
{\sum_{j=1}^N \exp \left(\text{sim}\left(\bm{t}_{(aug, i)}, \bm{v}_{(aug, j)}\right) / \tau_2\right)}.
\end{aligned}
\label{eq:instance_contrastive}
\end{equation}
The final contrastive loss is the average of the two, which is computed as:
\begin{equation}
\mathcal{L}_{\mathrm{ITC}_{aug}} = \frac{1}{2N}\sum_{i=1}^{N} (\ell_{(aug,i)}^{v2t} +\ell_{(aug,i)}^{t2v}).
\label{eq:contrastive_loss}
\end{equation}
Since MedCutMix operates as a plug-in and runs simultaneously with the base model's training, we reuse the original objectives of the base model~\cite{mgca_wang2022multi}, which include a token-wise alignment loss ($\mathcal{L}_{\mathrm{CTA}}$), instance-wise alignment loss ($\mathcal{L}_{\mathrm{ITA}}$), and a prototype-level alignment loss ($\mathcal{L}_{\mathrm{CPA}}$). Therefore, the final loss is expressed as follows:
\begin{equation}
\mathcal{L} = \mathcal{L}_{\mathrm{CTA}} + \mathcal{L}_{\mathrm{ITA}} + \mathcal{L}_{\mathrm{CPA}} + \mathcal{L}_{\mathrm{ITC}_{aug}}.
\end{equation}
MedCutMix strengthens contrastive pre-training by emphasizing disease-specific regions, enhancing the semantic diversity of the original dataset. Additionally, as disease-related image regions are guided by corresponding diagnostic reports, the CutMix areas between images and reports remain semantically aligned, ensuring both the coherence and quality of the augmented data.

\section{Experiments}
\label{sec:Experiments}

\subsection{Experimental Details} 
\noindent \textbf{Pre-training Setup.} 
For our pre-training phase, we utilize the MIMIC-CXR-JPG dataset~\cite{mimiccxr_johnson2019mimic}. Adhering to protocols established by previous research~\cite{mgca_wang2022multi}, we select only frontal-view chest images and extract the impression and findings from the accompanying radiological reports, yielding over 210,000 radiograph-report pairs. We follow the base framework~\cite{mgca_wang2022multi} and use the ViT-B/16~\cite{vit_dosovitskiy2020image} as the image encoder and employ BioClinicalBERT~\cite{bioclinicalbert_alsentzer2019publicly} as the text encoder. The batch size is set to 72 on a single GPU, while 2 GPUs are used for pre-training. MedCutMix is applied after 5 epochs of warm-up. In MedCutMix augmentation, we focus on common chest X-ray disease findings suggested by~\cite{cxp_irvin2019chexpert} for pairwise augmentation. We empirically set the sentence temperature $\mathbb{\tau}_1$ to 0.005, while the intermediate image features are extracted from the 11th layer of the ViT.

% \vspace{+0.2cm}
\noindent \textbf{Downstream Setup.} 
The efficacy of MedCutMix is validated through a comparative assessment of the transferability of baseline models with and without MedCutMix. Our evaluation focus on  
zero-shot image classification across various datasets, with results presented as the macro average of AUROC and F1 scores across all categories.
\textbf{CheXpert}~\cite{cxp_irvin2019chexpert} features 224,316 chest X-ray images from 65,240 patients, collected at Stanford Hospital. The official validation set consists of 200 chest radiographic studies annotated by three board-certified radiologists, while the official test set includes 500 studies annotated by five board-certified radiologists. Model selection is based on zero-shot AUROC performance on the validation set. During testing, results are reported both for the 5 observations in the official test set for competition tasks and for the full set of 14 disease labels.
\textbf{NIH}~\cite{nih_wang2017chestx} 
% consists of 112,120 frontal-view X-ray images sourced from 30,805 patients, gathered by the National Institutes of Health (NIH) between 1992 and 2015. The dataset includes labels for 14 common diseases. We adhered to the dataset splits available on . % https://www.kaggle.com/datasets/nih-chest-xrays/data
was utilized in a multi-label classification setup, comprising a total of 100k frontal-view X-ray images of about 32,000 patients annotated with 14 Chest X-Ray CXR diseases.
% (that is, atelectasis, cardiomegaly, consolidation, oedema, effusion, emphysema, fibrosis, hernia, infiltration, mass, nodule, pleural thickening, pneumonia and pneumothorax). 
We used offical NIH test set for evaluation.
\textbf{PadChest}~\cite{pdc_bustos2020padchest} comprises over 160,000 chest X-ray images from approximately 67,000 patients at Hospital San Juan (Spain). It has 193 disease image labels, including 174 radiographic 254 findings and 19 differential diagnoses. We adopt 39,053 chest X-rays annotated by board-certified 255 radiologists for zero-shot evaluation.
\textbf{RSNA Pneumonia}~\cite{rsna_shih2019augmenting} was used in its stage 2 version, followed~\cite{mgca_wang2022multi}. It comprises approximately 29,700 frontal-view chest radiographs. The task involves binary classification, distinguishing each chest image as either normal or pneumothorax positive. We employed the data split provided by~\cite{mgca_wang2022multi}.

\subsection{Experimental Results}
% We perform a comparative analysis of the downstream performance of our baselines ($i.e.,$ MGCA and GLoRIA) with and without the integration of MedCutMix. 
% Table~\ref{main_results} presents he zero-shot classification performance on the test sets. When either MGCA or GLoRIA serves as the base model, integrating MedCutMix results in enhanced average AUROC scores. This improvement is particularly pronounced within the MGCA framework, where AUROC sees an increase across all datasets with the application of MedCutMix. We hypothesize that this is due to MGCA's use of multi-granularity alignment objectives, which enhance cross-modal alignment, thereby allowing MedCutMix to perform more effectively.
% \input{tables/main_results}
\begin{figure*}[!t]
\centering
    \begin{center}
    \includegraphics[width=1.0\textwidth]{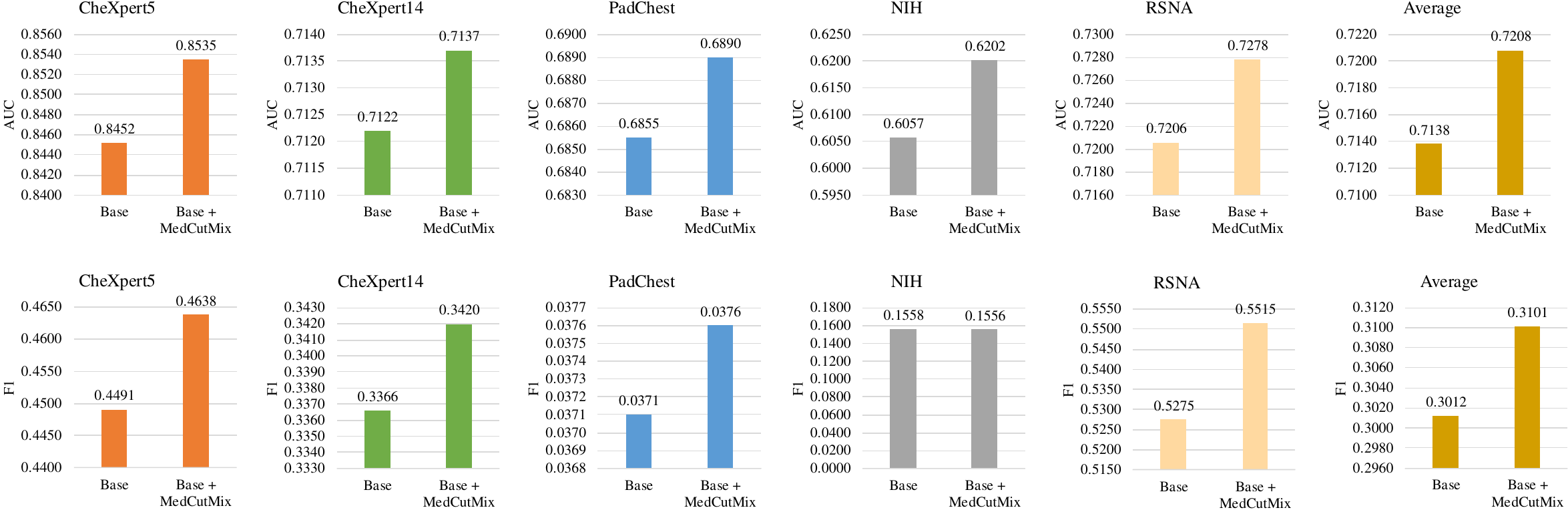}
    \end{center}
    \vspace{-20pt}
    \caption{Performance comparison of the base model without and with MedCutMix across multiple radiology datasets. The top row shows the improvement in AUC scores, while the bottom row presents F1-score gains.}
    \label{fig:auc_f1}
\end{figure*}
The results Fig.~\ref{fig:auc_f1} show the results of the base model without and with MedCutMix, which demonstrate the effectiveness of MedCutMix, a disease-centric data augmentation technique, in improving model performance across multiple radiology diagnosis tasks. Notably, MedCutMix enhances AUC scores, particularly in the CheXpert 5-Class task (from 0.8452 to 0.8535), while also yielding gains in CheXpert 14-Class (0.7122 to 0.7137), NIH (0.6057 to 0.6202), and RSNA (0.7206 to 0.7278), leading to an overall average AUC improvement from 0.7138 to 0.7208. Furthermore, F1 scores also see a significant boost, with CheXpert 5-Class increasing from 0.4491 to 0.4638 and RSNA improving from 0.5275 to 0.5515. These consistent gains highlight MedCutMix’s ability to generate more informative and diverse training samples, enhancing both model generalization and robustness.

\subsection{Discussions}
To evaluate the impact of hyper-parameter settings on MedCutMix, we explored the effects of varying the maximum number of mixed samples ($N_{max}$) and the mixed layer ($k$). 

\noindent \textbf{Analysis of the number of mixed samples.}
% Table~\ref{sample_num} illustrates the zero-shot AUC performances of the MGCA base model with MedCutMix using different maximum numbers of mixed samples while maintaining the mix layer in the 11th layer. The performance is optimal when the maximum number of mixed samples is set to 40. When the number of mixed samples is relatively low, such as 30, the augmentation effect is limited, and the performance approximates the baseline. However, on the other extreme, if the number of mixed samples becomes excessive, like 300, and considering our batch size of 72, it leads to a scenario where there is more augmented data than real data. This may disrupt the trade-off between introducing noise and preserving diversity.
The results in Table~\ref{sample_num} highlight the impact of varying $N_{max}$ on model performance, revealing a balance between augmentation benefits and potential drawbacks. The highest average AUC (0.7208) is achieved at $N_{max}$ = 40, demonstrating that moderate augmentation enhances generalization by introducing meaningful diversity while preserving essential clinical features. However, as $N_{max}$ increases to 300, performance declines significantly (AUC drops to 0.7054), likely due to an overabundance of synthetic samples disrupting the real data distribution and reducing feature fidelity. Dataset-specific effects are also observed, where different datasets exhibit varying sensitivity to synthetic augmentation. These findings underscore that while moderate augmentation strengthens model robustness, excessive artificial mixing may degrade clinical relevance and hinder learning.

\noindent \textbf{Analysis of the number of mixed layers.} Table~\ref{layer} presents the impact of varying the intermediate layers ($k$) at which visual features are mixed, with the number of mixed samples held constant. Specifically, we observed highest performance is achieved when mixing occurs at the final layer and diminishes when applied to earlier layers. Nonetheless, all variations still outperform the baseline, which does not employ MedCutMix.

\input{tables/sample_num}
\input{tables/layer}

\section{Conclusion}
This paper presents MedCutMix, a novel multi-modal disease-centric data augmentation technique specifically tailored to address data scarcity in VLP within the radiology domain. Our approach performs CutMix of diagnostic sentences within reports and mix feature-level representations of disease-attended images. The method effectively enhances the semantic diversity of augmented medical data while preserving cross-modal semantic coherence, thus addressing the challenges posed by privacy concerns and labeling costs. Our comprehensive evaluation demonstrates that MedCutMix outperforms the advanced medical VLP baseline in four datasets. These results affirm the efficacy of MedCutMix in improving the performance and generalizability of models within the medical VLP.

\vfill\pagebreak

\bibliographystyle{IEEEbib}
% \bibliography{strings,refs}
\bibliography{main}

\end{document}

%% file: tables/sample_num.tex
\begin{table}[t!]
\centering
% \vspace{-5pt}
\caption{AUC scores under different maximum MedCutMix samples ($N_{max}$). 
% The best performance is achieved at $N_{max}$ = 40, while excessive mixing ($N_{max}$ = 300) negatively impacts performance.
}
% \vspace{-5pt}
\label{sample_num}
\setlength\tabcolsep{4.0pt}
\resizebox{0.5\textwidth}{!}{
\begin{tabular}{c|c|c|c|c|c|c}
\toprule[1.5pt]
\multirow{2}{*}{$N_{max}$} & \multicolumn{5}{c|}{Datasets} & \multirow{2}{*}{Avg. AUC} \\ 
& CheXpert5 & CheXpert14 & PadChest & NIH & RSNA & \\
\hline
0                                                        & 0.8452    & 0.7122     & 0.6855   & 0.6057           & 0.7206 & 0.7138          \\ \hline
30                                                       & 0.8492    & 0.7158     & \textbf{0.7011}   & 0.6196           & 0.6412 & 0.7054          \\ \hline
40                                                       & 0.8535    & 0.7137     & 0.6890    & \textbf{0.6202}           & 0.7278  & \textbf{0.7208} \\ \hline
50                                                       & 0.8534    & 0.7137     & 0.6890    & \textbf{0.6202}           & 0.7276 & 0.7208          \\ \hline
100                                                      & \textbf{0.8574}     & \textbf{0.7415}      & 0.6960    & 0.6183           & 0.6097 & 0.7046          \\ \hline
300                                                      & 0.8360     & 0.6768     & 0.6781   & 0.6068           & \textbf{0.7294} & 0.7054          \\ \bottomrule[1.5pt]

\end{tabular}
}
\end{table}

%% file: tables/layer.tex
\begin{table}[t!]
\centering
% \vspace{-2pt}
\caption{AUC scores under different mixing layers ($k$). 
% The highest performance is obtained when applying MedCutMix at the final layer ($k$ = 11), while earlier layers exhibit a slight drop in AUC.
}
% \vspace{-2pt}
\label{layer}
\setlength\tabcolsep{4.0pt}
\resizebox{0.5\textwidth}{!}{
\begin{tabular}{c|c|c|c|c|c|c}
\toprule[1.5pt]
\multirow{2}{*}{Layer $k$} & \multicolumn{5}{c|}{Datasets} & \multirow{2}{*}{Avg. AUC} \\ 
% \cdashline{2-6}
& CheXpert5 & CheXpert14 & PadChest & NIH & RSNA & \\
\hline
5                                               & 0.8528    & 0.7100       & 0.6912   & 0.6206           & 0.7230  & 0.7195  \\ \hline
8                                               & 0.8532    & 0.7130      & 0.6891   & 0.6201           & 0.7272 & 0.7205  \\ \hline
11                                              & 0.8534    & 0.7137     & 0.6890    & 0.6202           & 0.7276 & \textbf{0.7208}  \\ \bottomrule[1.5pt]
\end{tabular}
}
\end{table}